\documentclass[10pt,twocolumn,letterpaper]{article}

\usepackage[pagenumbers]{cvpr}      % To produce the REVIEW version
\newcommand{\figref}[1]{Fig.~\ref{#1}}
\newcommand{\secref}[1]{Section~\ref{#1}}

\newcommand{\tabref}[1]{Table~\ref{#1}}

\usepackage[table]{xcolor}
\definecolor{cvprblue}{rgb}{0.21,0.49,0.74}
\usepackage[pagebackref,breaklinks,colorlinks,citecolor=cvprblue]{hyperref}
\usepackage{graphicx}
\usepackage{amsmath}
\usepackage{amssymb}
\usepackage{booktabs}
\usepackage{bm}
\usepackage{tikz}
\usepackage{graphicx}
\usepackage{booktabs}
\usepackage{etoolbox}
\usepackage[tracking=true]{microtype}
\usepackage{multirow}
\usepackage{enumitem}
\usepackage{comment}
\usepackage{float}
\usepackage{lipsum}

\usetikzlibrary{calc}

\DeclareMathOperator*{\argmax}{arg\,max}

\def\PSNRColName{\textls[-20]{PSNR\kern-0.05em$\uparrow$}}
\def\SSIMColName{\textls[-20]{SSIM\kern-0.05em$\uparrow$}}
\def\LPIPSColName{\textls[-40]{LPIPS\kern-0.05em$\downarrow$}}
\def\AlgorithmColName{{\textbf{Algorithm\kern-0.05em}}}

\usepackage[capitalize]{cleveref}

\newcommand{\boldparagraph}[1]{\noindent{\bf #1} }

\def\confName{3DV\xspace}
\def\confYear{2025\xspace}

\title{AGS-Mesh: Adaptive Gaussian Splatting and Meshing with Geometric Priors \\ for Indoor Room Reconstruction Using Smartphones}

\author{
Xuqian Ren${}^{1 \ddagger}$ \and Matias Turkulainen${}^2$ \and Jiepeng Wang${}^{3 \dagger}$ \and
Otto Seiskari${}^4$ \and Iaroslav Melekhov${}^2$ \hspace{0.5cm} 
Juho Kannala${}^{2,5}$ \hspace{0.5cm} Esa Rahtu${}^1$ \\
$^1$~Tampere University, $^2$~Aalto University,
$^3$~University of Hong Kong, \\ $^4$~Spectacular AI, $^5$~University of Oulu
}

\begin{document}
\twocolumn[{
\renewcommand\twocolumn[1][]{#1}
\maketitle
\vspace{-3em}
\begin{center}
        \captionsetup{type=figure}
        \centering
        \hspace{-50pt}     
        \begin{subfigure}[b]{0.18\textwidth}
            \centering
            \begin{tikzpicture}
                \node [inner sep=0pt,outer sep=0pt,clip,rounded corners=2pt] at (0,0) {\includegraphics[height=4.7cm]{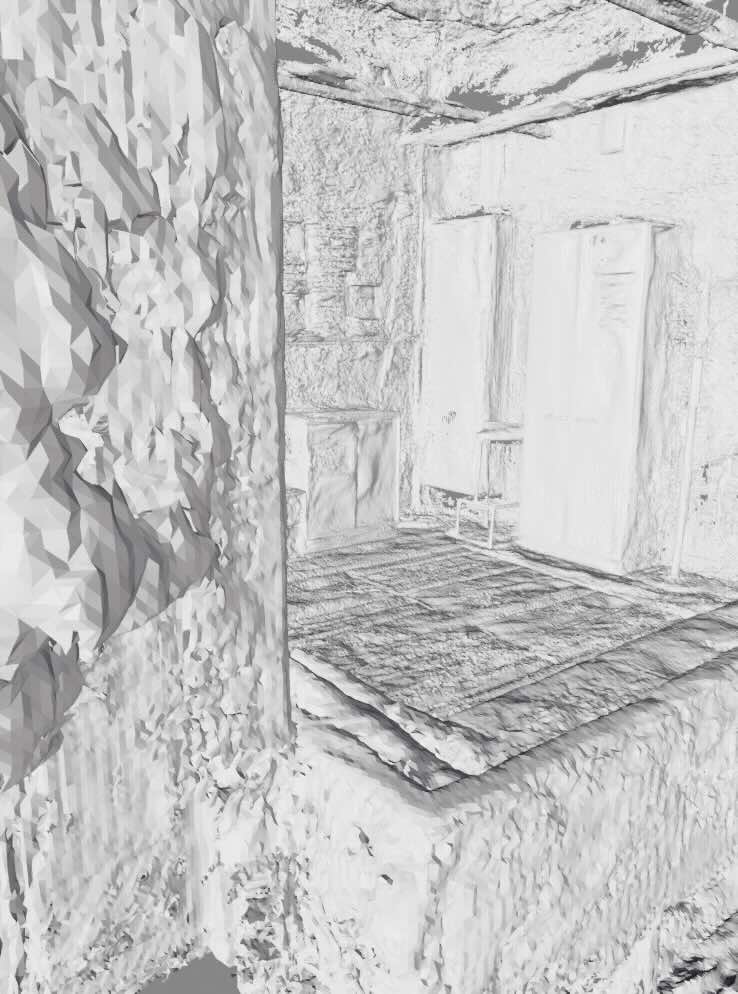}};
                \node[font=\tiny, text=white] at (0.2,-1.1) {};
            \end{tikzpicture}
            \subcaption{3DGS~\cite{kerbl20233d}}
        \end{subfigure}
         \hspace{8pt}
        \begin{subfigure}[b]{0.18\textwidth}
            \centering
            \begin{tikzpicture}
                \node [inner sep=0pt,outer sep=0pt,clip,rounded corners=2pt] at (0,0) {\includegraphics[height=4.7cm]{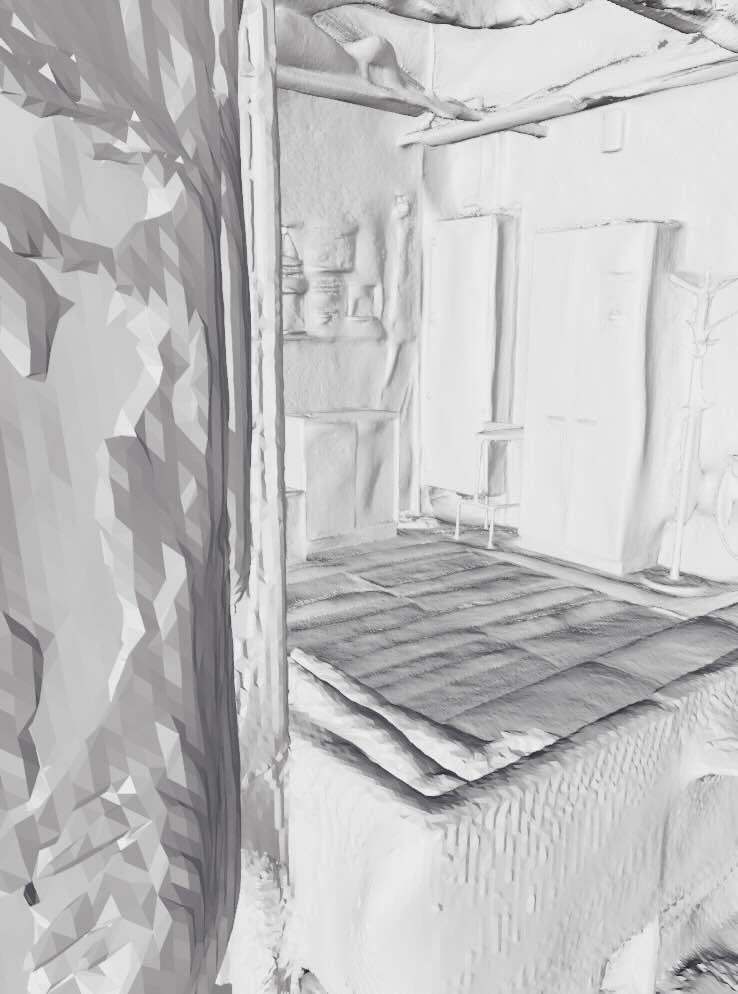}};
                \node[font=\tiny, text=white] at (0.2,-1.1) {};
            \end{tikzpicture}
            \subcaption{2DGS~\cite{Huang2DGS2024}}
        \end{subfigure}
        \hspace{8pt}
        \begin{subfigure}[b]{0.18\textwidth}
            \centering
            \begin{tikzpicture}
                \node [inner sep=0pt,outer sep=0pt,clip,rounded corners=2pt] at (0,0) {\includegraphics[height=4.7cm]{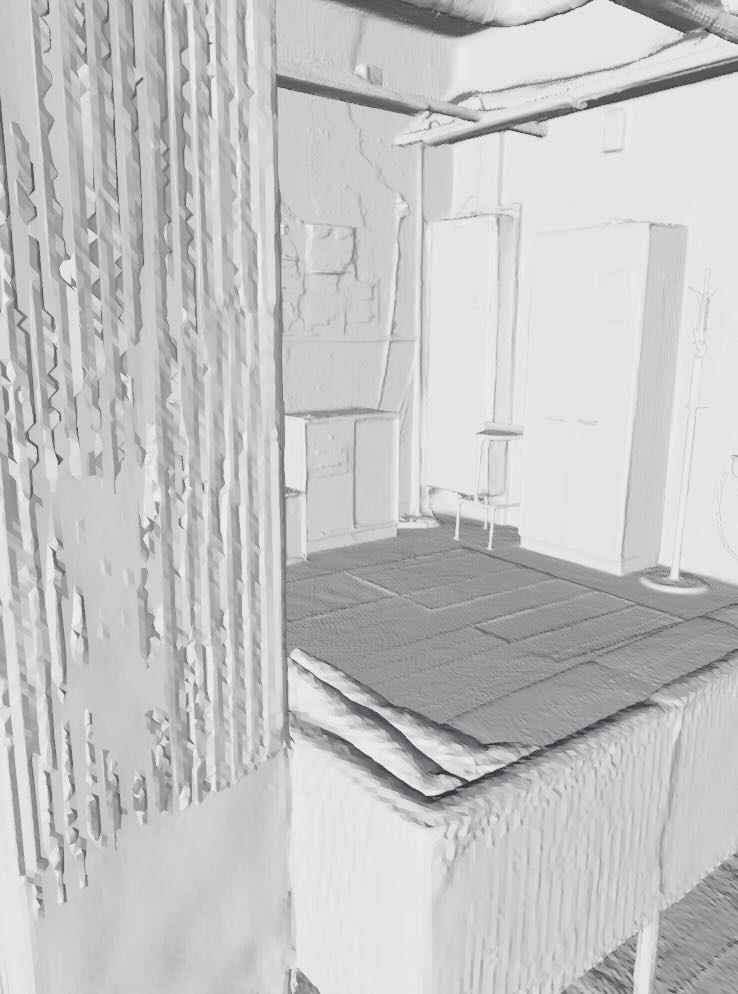}};
                \node[font=\tiny, text=white] at (0.2,-1.1) {};
            \end{tikzpicture}
            \subcaption{DN-Splatter~\cite{turkulainen2024dn}}
        \end{subfigure}
        \hspace{8pt}
        \begin{subfigure}[b]{0.18\textwidth}
            \centering
            \begin{tikzpicture}
                \node [inner sep=0pt,outer sep=0pt,clip,rounded corners=2pt] at (0,0) {\includegraphics[height=4.7cm]{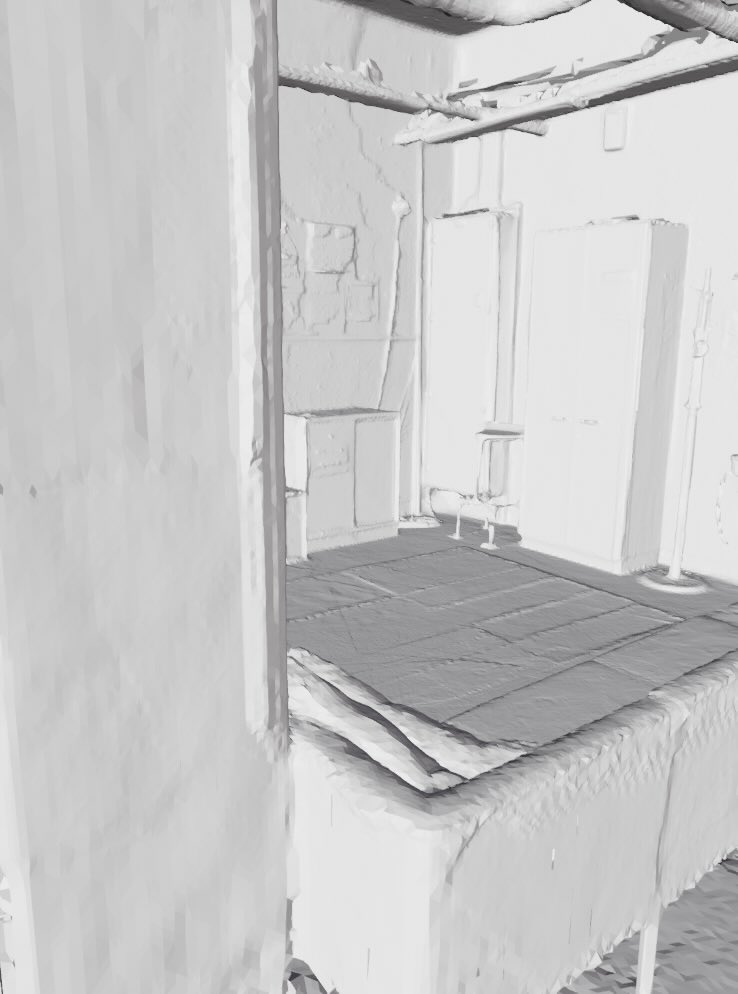}};
                \node[font=\tiny, text=white] at (0.2,-1.1) {};
            \end{tikzpicture}
            \subcaption{2DGS + Ours}
        \end{subfigure}
        \hspace{8pt}
        \begin{subfigure}[b]{0.18\textwidth}
            \centering
            \begin{tikzpicture}
                \node [inner sep=0pt,outer sep=0pt,clip,rounded corners=2pt] at (0,0) {\includegraphics[height=4.7cm]{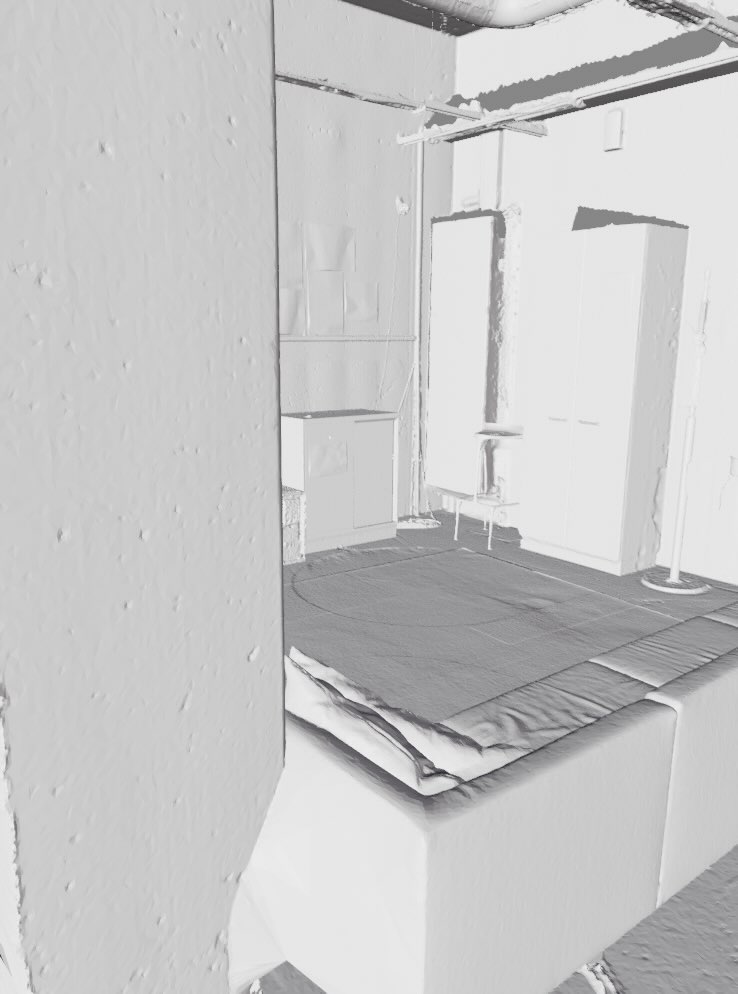}};
                \node[font=\tiny, text=white] at (0.2,-1.1) {};
            \end{tikzpicture}
            \subcaption{Reference mesh}
        \end{subfigure}
        \hspace{-40pt}
\caption{We present AGS-Mesh, a method that adaptively integrates geometric priors into Gaussian Splatting for indoor room reconstruction using a mobile device. We enhance existing Gaussian Splatting based methods (a), (b), (c) by filtering inconsistent prior estimates during optimization and utilize an IsoOctree based meshing strategy that recovers a smoother scene surface with higher detail (d).}
\end{center}
}]

\renewcommand{\thefootnote}{$\ddagger$}
\footnotetext[1]{Denotes Project Lead}
\renewcommand{\thefootnote}{$\dagger$}
\footnotetext[2]{Denotes Corresponding Author}

\begin{abstract}
Geometric priors are often used to enhance 3D reconstruction. With many smartphones featuring low-resolution depth sensors and the prevalence of off-the-shelf monocular geometry estimators, incorporating geometric priors as regularization signals has become common in 3D vision tasks. However, the accuracy of depth estimates from mobile devices is typically poor for highly detailed geometry, and monocular estimators often suffer from poor multi-view consistency and precision. In this work, we propose an approach for joint surface depth and normal refinement of Gaussian Splatting methods for accurate 3D reconstruction of indoor scenes. We develop supervision strategies that adaptively filters low-quality depth and normal estimates by comparing the consistency of the priors during optimization. We mitigate regularization in regions where prior estimates have high uncertainty or ambiguities. Our filtering strategy and optimization design demonstrate significant improvements in both mesh estimation and novel-view synthesis for both 3D and 2D Gaussian Splatting-based methods on challenging indoor room datasets. Furthermore, we explore the use of alternative meshing strategies for finer geometry extraction. We develop a scale-aware meshing strategy inspired by TSDF and octree-based isosurface extraction, which recovers finer details from Gaussian models compared to other commonly used open-source meshing tools. Our code is released in \url{https://xuqianren.github.io/ags_mesh_website/}.

\end{abstract}
\vspace{-5mm}
\section{Introduction}
\label{sec:intro}
% story draft
Photorealistic and geometrically accurate reconstruction of real-world indoor scenes has various applications in virtual reality, augmented reality, and video games. Traditional approaches have addressed the problem by creating textured meshes that can be rendered using conventional graphics pipelines. Depth sensors, such as high-precision 3D LiDAR scanners or Kinect sensors, are often used to aid geometric reconstruction; however, these devices are generally expensive for consumer users and require considerable technical expertise. With the rapid development of consumer mobile devices, the latest iPhone smartphones now come equipped with multiple high-resolution RGB cameras and miniaturized LiDAR sensors, making them ideal tools for millions of users to collect 3D data. While the low-resolution depth maps obtained from such LiDAR sensors enable accurate planar reconstruction, they struggle with highly detailed geometry and edges, as shown in~\figref{fig:cmp_iphone_kinect_depth}. These inaccuracies become evident when using the low-resolution depth maps for 3D mesh reconstruction with traditional methods like volumetric fusion \cite{Izadi2011KinectFusionR3, curless1996volumetric}.

Recent advances in differentiable inverse rendering have provided alternative 3D representations to meshes, achieving photorealistic novel-view synthesis with relatively fast rendering and training times. 3D Gaussian Splatting (3DGS)-based methods~\cite{kerbl20233d, turkulainen2024dn, guedon2023sugar, xiong2023sparsegs, yu2024gaussian} represent 3D scenes with millions of differentiable 3D Gaussians, enabling high-quality and real-time rendering. However, the geometry extracted from such scenes for larger scenes is poor. 2D Gaussian Splatting (2DGS)~\cite{Huang2DGS2024} further extends 3DGS towards mesh reconstruction, showing promising results for object-centric reconstruction. However, performance on room-scale reconstruction with data captured by a mobile device is still lacking. Low-texture surfaces and sparse, outward-facing captures, common in indoor room datasets \cite{yeshwanthliu2023scannetpp, ren2023mushroom}, pose challenges and ambiguities for purely photometric-based reconstruction. Given these observations, it is clear that achieving photorealistic and geometrically accurate reconstructions of common indoor scenes using consumer devices remains an open challenge.

In this work, we extend recent state-of-the-art Gaussian Splatting approaches~\cite{turkulainen2024dn, Huang2DGS2024} by adaptively combining low-resolution depth estimates and off-the-shelf monocular geometry estimates for high-fidelity indoor room reconstruction using a mobile device. Specifically, we design two regularization strategies to address inconsistencies between low-resolution depth maps and off-the-shelf monocular normal estimates during training. We regularize Gaussian positions using low-resolution depth maps on planar and smooth surfaces. We introduce a new depth regularization strategy, coined Depth Normal Consistency (DNC), to filter inconsistencies in depth maps by considering the consistency between normals derived from noisy depth maps and those estimated via pretrained monocular networks. Additionally, we propose an Adaptive Normal Regularization strategy (ANR) to refine normals by mitigating regularization in regions where monocular normal estimators struggle to provide accurate prior estimates.

By carefully filtering inconsistencies between various geometric estimates during training, our approach improves both novel-view synthesis and geometry extraction compared to previous methods. Furthermore, we develop a meshing strategy inspired by TSDF~\cite{curless1996volumetric, Izadi2011KinectFusionR3} and octree-based isosurface extraction~\cite{isooctree} methods, which accounts for the hierarchical details present in larger indoor scenes. This post-processing strategy enhances the fine details recovered from the 3D Gaussian scene without requiring additional optimization, surpassing conventional Poisson and Marching Cubes-based~\cite{lorensen1998marching} TSDF alternatives commonly used for 3D mesh reconstruction.

We summarize our contributions with the following statements:
\begin{itemize}[leftmargin=*]
    \item We propose a novel regularization strategy for indoor room reconstruction that adaptively filters geometric priors from mobile devices and off-the-shelf monocular estimators, enhancing photorealism and geometry reconstruction from Gaussian Splatting-based methods.
    \item We introduce a mesh post-processing method based on adaptive TSDF and IsoOctree meshing that recovers finer scene details compared to commonly used alternatives.
    \item We demonstrate, through extensive experiments, improvements in both novel-view synthesis and geometry extraction for high-fidelity indoor room reconstruction.
\end{itemize}

\section{Related Work}
\label{sec:related_work}
\boldparagraph{Traditional Meshing Techniques.}
Classical 3D reconstruction methods utilize multi-view stereo (MVS) techniques. Prior methods use RGB images as input~\cite{goesele2007multi,goesele2007multi,vu2011high,ding2022transmvsnet}, while some integrate depth maps~\cite{henry2012rgb,curless1996volumetric,merrell2007real,prisacariu2017infinitam} for 3D reconstruction. Learning-based methods~\cite{yu2020fast,weder2021neuralfusion,riegler2017octnetfusion,patchM} extract, match, and fuse image features by utilizing neural networks, often improving the quality of reconstruction compared to classical methods. However, they struggle to accurately reconstruct casually captured scenes with poor textures, which are common in everyday scenes.

\boldparagraph{Meshing With Neural Implicit Representations.}
Recent neural implicit representations, particularly NeRF-based methods~\cite{mildenhall2020nerf,barron2022mipnerf,M_ller_2022}, excel at novel-view synthesis by encoding 3D scenes into multi-layer perceptrons (MLPs) and using volume rendering~\cite{Kajiya} to synthesize views. However, these methods struggle with accurate geometry and physical surfaces and are primarily for photorealistic rendering. Some approaches~\cite{roessle2022dense,deng2022depth,wei2021nerfingmvs} incorporate depth supervision for better geometry, and others~\cite{Rakotosaona2023THREEDV,tang2022nerf2mesh} attempted to extract watertight meshes from NeRFs. However, these methods are generally limited to small synthetic objects or carefully constructed inward-facing captures.

Alternatively, Signed Distance Function (SDF)-based methods~\cite{wang2021neus,yariv2021volume} define surfaces as level set crossings of an implicit function. These methods produce smoother surfaces with Marching Cubes~\cite{lorensen1998marching} but often lose detail in thin objects, perform worse in novel-view synthesis, and are costly to train on consumer GPUs.

\boldparagraph{Utilizing Geometric Priors for Meshing.}
MonoSDF~\cite{Yu2022MonoSDF} uses depth and normal predictions from a pretrained model as regularization signals to improve an SDF-based implicit model \cite{wang2021neus}. However, monocular priors can yield inconsistent predictions, with poor multi-view consistency, leading to misguided regularization signals during optimization. Neural RGB-D~\cite{azinovic2022neural} and GO-Surf~\cite{wang2022go} address this by using real sensor depth readings for indoor room reconstruction. Other methods~\cite{wang2022neuris,yu2023improving} adaptively apply normal priors, while DebSDF~\cite{xiao2024debsdf} and H$_2$O-SDF~\cite{park2024h2osdf} manage uncertainty in priors. Choi et al.~\cite{choi2023tmo} also observe that noisy depth readings can hinder training and attempt to filter them as a pre-processing step. In this work, we adaptively mitigate inconsistencies in priors during optimization of Gaussian scenes to achieve better geometric reconstructions.

\boldparagraph{Meshing Using Gaussian Splatting.}
3D Gaussian Splatting (3DGS)~\cite{kerbl20233d} significantly improves novel-view synthesis speed by direct projection and blending of Gaussian points instead of the more costly ray tracing method in NeRF. To obtain a mesh from 3DGS, follow up work have employed different techniques: optimizing 3D Gaussians along with mesh optimization methods~\cite{guedon2023sugar} or by replacing 3D Gaussians with 2D surfels~\cite{Huang2DGS2024, dai2024high} to better align Gaussians with the surface. GOF~\cite{yu2024gaussian} learns a Gaussian opacity field and directly extracts a mesh from it using tetrahedral grids. %NeuSG~\cite{chen2023neusg} simultaneously learns both a SDF and Gaussian representation. 
Although these methods show promising performance on object-level data, realistic indoor room reconstruction remains challenging. DN-Splatter~\cite{turkulainen2024dn} demonstrated that sensor depth and monocular normals can help in extracting meshes from larger indoor scenes. However, noisy sensor depth from mobile devices and inconsitencies in priors can harm optimization. VCR-GauS~\cite{chen2024vcr} addresses this by learning a confidence term to weigh normal regularization. In our work, we focus on achieving 3D room reconstruction by adaptively integrating noisy geometric priors into Gaussian Splatting frameworks. This approach facilitates real-time photorealistic image rendering, accelerates training for mesh generation, and more effectively utilizes geometric priors.

\boldparagraph{Mesh Extraction Algorithms.} We briefly summarize post-processing techniques for mesh extraction from 3D representations in the closest prior work. To extract a mesh from a Gaussian scene, 2DGS~\cite{Huang2DGS2024} uses a well-established TSDF fusion algorithm from~\cite{Zhou2018}. SuGaR~\cite{guedon2023sugar} utilizes Poisson reconstruction~\cite{kazhdan2006poisson} to reconstruct a watertight mesh from an oriented point cloud. GOF~\cite{yu2024gaussian} establishes tetrahedral grids around each Gaussian and applies Marching Tetrahedra~\cite{shen2021deep} to extract a triangle mesh. In this work, we modify the classical TSDF with a scale-aware aspect and replace the naive Marching Cubes with the unconstrained octree isosurface extraction method~\cite{kazhdan2007unconstrained} which we refer to as \emph{IsoOctree} in this work. This allows adapting the level of detail in the mesh to the varying precision in the scene, producing more optimized meshes with less computation.

%\vspace{-0.5cm}
\section{Preliminaries}
\label{sec:preliminaries}
\subsection{Geometric Priors from Handheld Devices}
\begin{figure}[t!]
\centering
\vspace{-0.2cm}
  \hspace{-70pt}
  \begin{minipage}[b]{0.2\textwidth} % Adjust width as needed
        \centering
        \begin{subfigure}[b]{0.45\textwidth}
            \centering
            \subcaption*{RGB}
            \begin{tikzpicture}
                \node [inner sep=0pt,outer sep=0pt,clip,rounded corners=2pt] at (0,0) {\includegraphics[height=2.5cm]{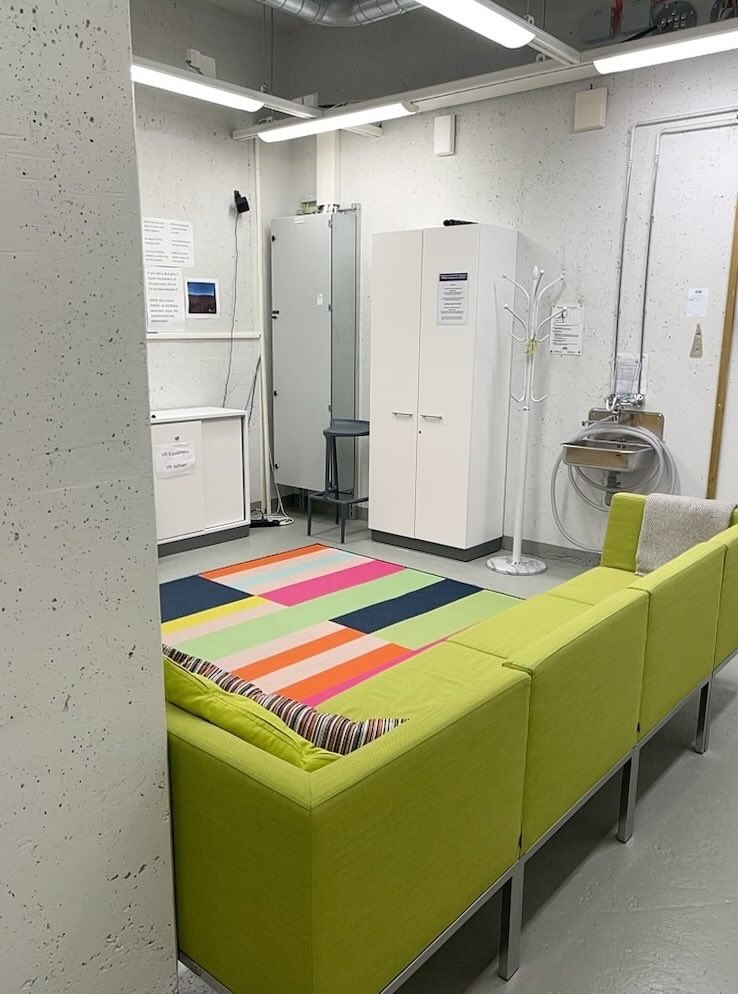}};
                \node[font=\tiny, text=white] at (0.2,-1.1) {};
                \draw[red,thick] (-0.3,-0.2) rectangle (0.1,0.8);
            \end{tikzpicture}
        \end{subfigure}
        \hspace{4pt}
        \begin{subfigure}[b]{0.45\textwidth}
            \centering
            \subcaption*{iPhone Depth}
            \begin{tikzpicture}
                \node [inner sep=0pt,outer sep=0pt,clip,rounded corners=2pt] at (0,0) {\includegraphics[height=2.5cm]{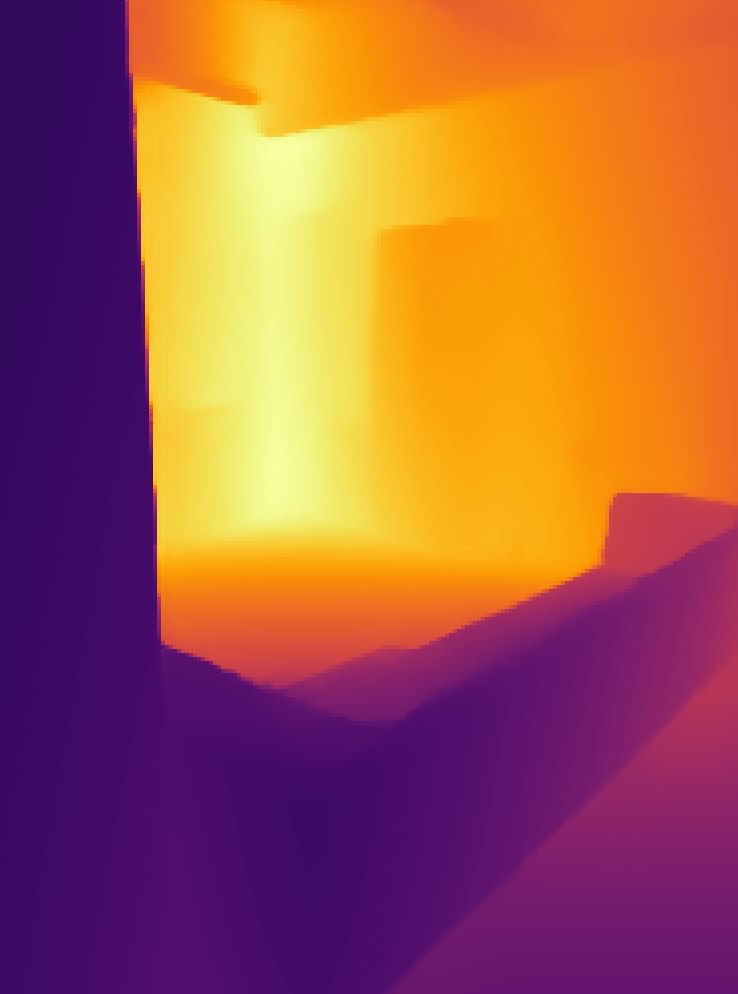}};
                \node[font=\tiny, text=white] at (0.2,-1.1) {};
                \draw[red,thick] (-0.3,-0.2) rectangle (0.1,0.8); 
            \end{tikzpicture}
        \end{subfigure}
        \centering
        \subcaption{}
    \end{minipage}
  \hspace{6pt}
  \begin{minipage}[b]{0.2\textwidth} % Adjust width as needed
        \centering
        \begin{subfigure}[b]{0.45\textwidth}
            \centering
            \subcaption*{RGB}
            \begin{tikzpicture}
                \node [inner sep=0pt,outer sep=0pt,clip,rounded corners=2pt] at (0,0) {\includegraphics[height=2.5cm]{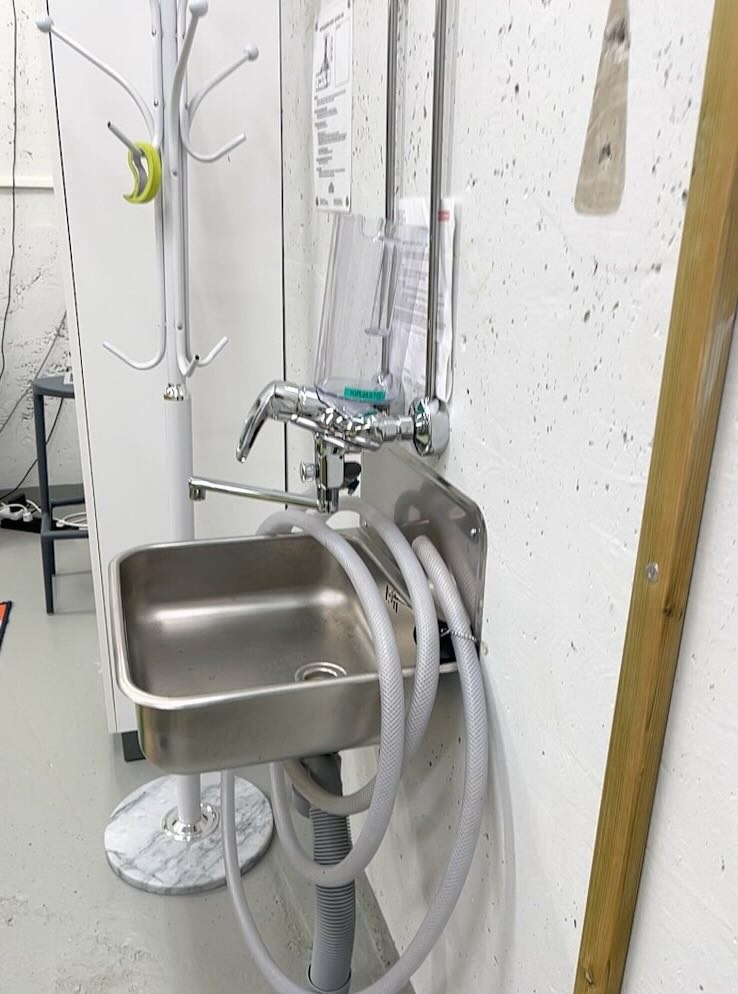}};
                \node[font=\tiny, text=white] at (0.2,-1.1) {};
                \draw[red,thick] (-0.9,-0.5) rectangle (-0.2,-1.2);
                \draw[red,thick] (-0.7,0.0) rectangle (-0.1,0.5);
            \end{tikzpicture}
        \end{subfigure}
        \hspace{4pt}
        \begin{subfigure}[b]{0.45\textwidth}
            \centering
            \subcaption*{iPhone Depth}
            \begin{tikzpicture}
                \node [inner sep=0pt,outer sep=0pt,clip,rounded corners=2pt] at (0,0) {\includegraphics[height=2.5cm]{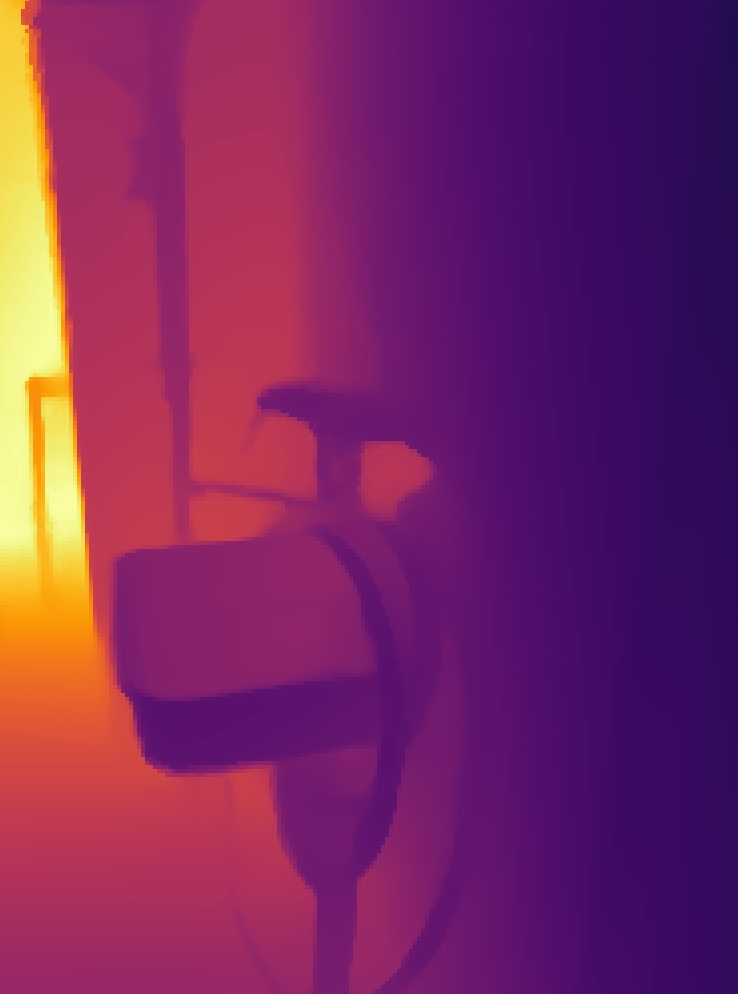}};
                \node[font=\tiny, text=white] at (0.2,-1.1) {};
                \draw[red,thick] (-0.9,-0.5) rectangle (-0.2,-1.2);
                \draw[red,thick] (-0.7,0.0) rectangle (-0.1,0.5);
            \end{tikzpicture}
        \end{subfigure}
        \centering
        \subcaption{}
    \end{minipage}
  \hspace{-70pt}

\hspace{-50pt}
\begin{minipage}[b]{0.5\textwidth} % Adjust width as needed
        \centering
        \begin{subfigure}[b]{0.45\textwidth}
            \centering
            \subcaption*{RGB}
            \begin{tikzpicture}
                \node [inner sep=0pt, outer sep=0pt, clip, rounded corners=2pt] at (0,0) {\includegraphics[height=2cm]{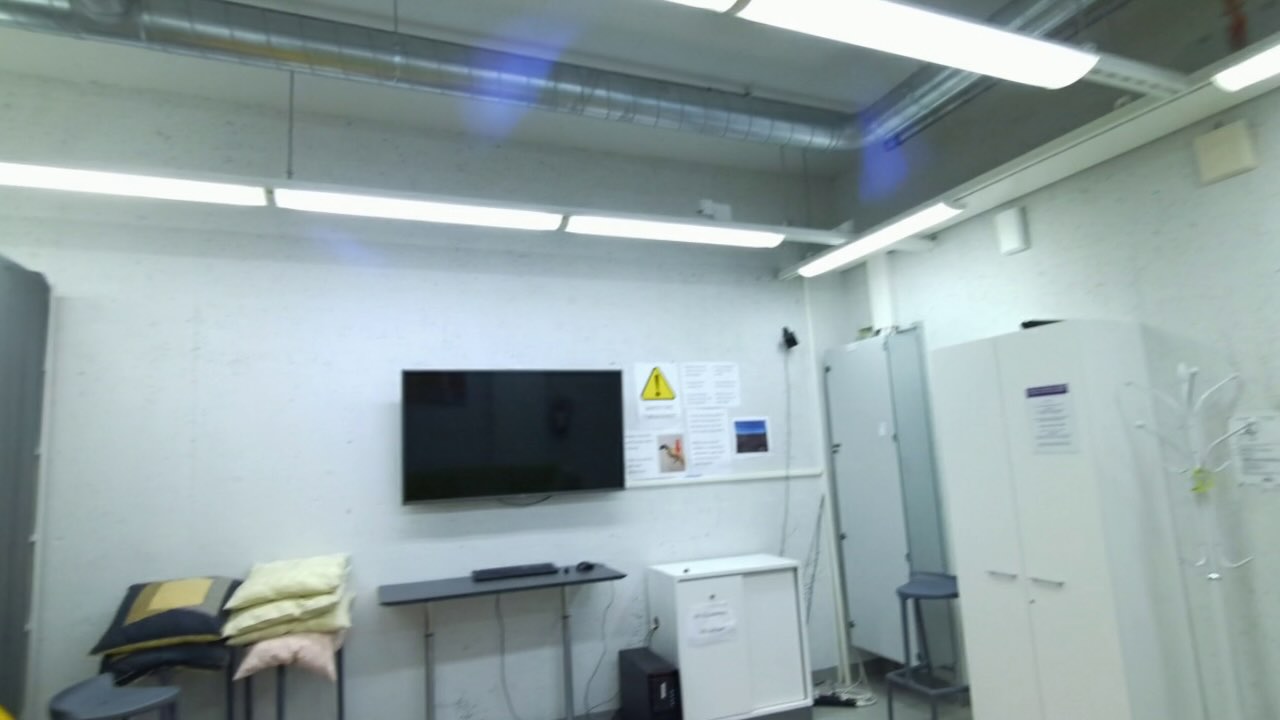}};
            \end{tikzpicture}
        \end{subfigure}
        \hspace{-10pt}
        \begin{subfigure}[b]{0.45\textwidth}
            \centering
            \subcaption*{Kinect Depth}
            \begin{tikzpicture}
                \node [inner sep=0pt, outer sep=0pt, clip, rounded corners=2pt] at (0,0) {\includegraphics[height=2cm]{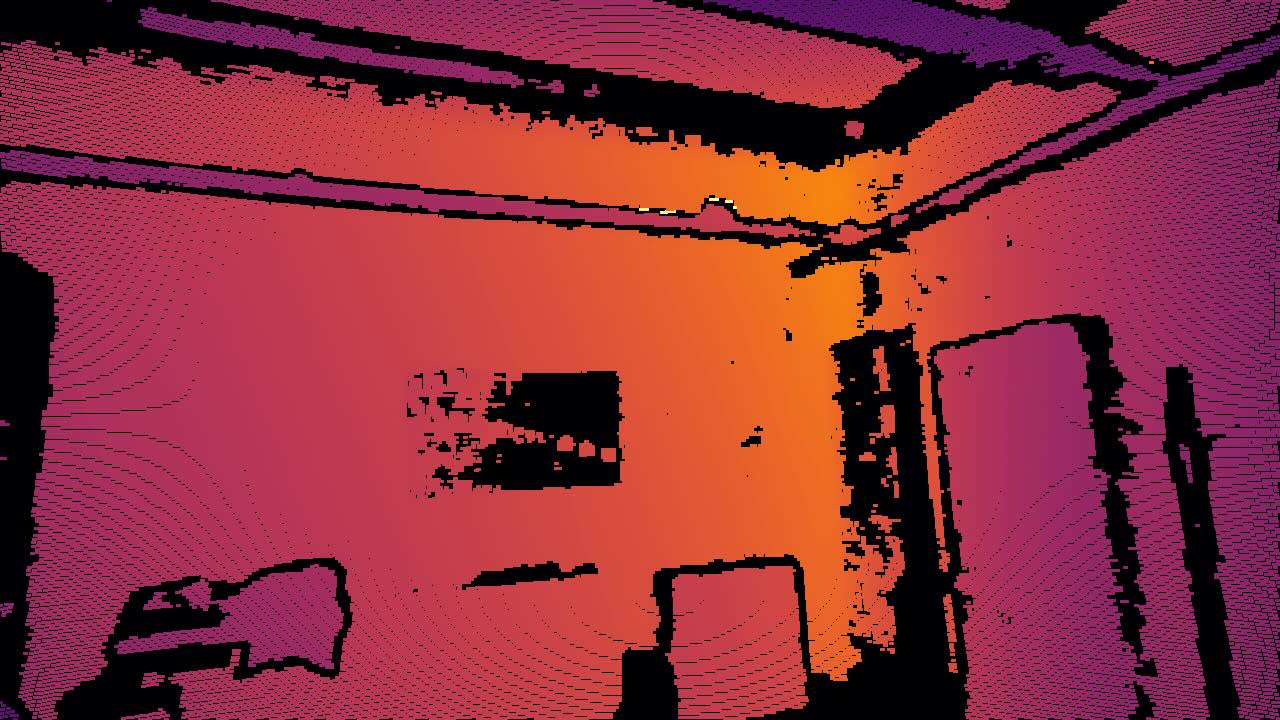}};
            \end{tikzpicture}
        \end{subfigure}
        \vspace{0.2cm}
        \centering
        \subcaption{}
    \end{minipage}
\hspace{-60pt}
  \caption{\textbf{Demonstration of iPhone and Kinect sensor depths.} The iPhone struggles to capture accurate depth values for (a) objects at a far distance, and (b) small objects, and edges. Instead, the Kinect sensor~\cite{microsoft-kinect-depth-camera} (c) is able to filter unconfident depth values} %for the same captured scene.}
   \vspace{-0.6cm}
\label{fig:cmp_iphone_kinect_depth}
\end{figure}

\boldparagraph{Depth from Kinect.}
Azure Kinect devices are popular in industry and research for indoor 3D reconstruction. They generate depth maps from a 1-megapixel Time-of-Flight (ToF) sensor with a $1024\times1024$ resolution with a per-pixel error range from 1.4mm to 12mm~\cite{khoshelham2012accuracy}.
Depths are denoised through post-processing by considering systematic and random errors. \figref{fig:cmp_iphone_kinect_depth} shows how the Kinect system adequately filters depth estimates for problematic regions that can occur, such as pixels being outside of the ToF illumination mask, saturated IR signal regions, and depth ambiguities due to poor sensor readings~\cite{microsoft-kinect-depth-camera}.

\boldparagraph{Depth From a Mobile Phone.}
Some of the latest smartphones, such as the iPhone Pro versions, are integrated with small LiDAR sensors that measure distances to objects. In the case of the iPhone, the resolution of the physical sensors is usually very small (\eg, only $16\times16$); however, sophisticated post-processing and enhancement with RGB frames are used to return depth maps with $256\times192$ resolution, synchronized with high-quality RGB images through developer APIs. Compared to Kinect, the depth estimates are low-resolution with an error range that can reach $\pm3$~cm horizontally and $\pm7$~mm vertically~\cite{chase2022apple}.
We illustrate the quality of the depth estimates in \figref{fig:cmp_iphone_kinect_depth}. We note that the low-resolution depths perform generally well for closeby objects but lose accuracy for faraway and very thin structures. 

\boldparagraph{Depth and Normal Priors from Monocular Networks.}
Recent monocular models~\cite{bhat2023zoedepth,depthanything,hu2024metric3dv2} can produce high-resolution metric depths and normal estimates from a single RGB image.
Although these models learn from large-scale image-geometry pairs, they struggle to achieve the same accuracy in metric depth compared to physical sensors found in devices like the iPhone or Kinect. Therefore, in this work, we utilize sensor depth measurements from mobile devices for indoor room reconstruction to obtain metrically accurate reconstruction.

\subsection{Gaussian Splatting}
3D Gaussian Splatting, introduced in~\cite{kerbl20233d}, explicitly represents a 3D scene with three-dimensional Gaussian primitives characterized by a center $\boldsymbol{\mu}$ and a covariance matrix $\boldsymbol{\Sigma}$:
\begin{equation}
G(\boldsymbol{x})=e^{-\frac{1}{2}(\boldsymbol{x}-\boldsymbol{\mu})^T \boldsymbol{\Sigma}^{-1}(\boldsymbol{x}-\boldsymbol{\mu})}
\end{equation}
where the covariance $\boldsymbol{\Sigma}$ is composed of scaling $\boldsymbol{S}$ and rotation $\boldsymbol{R}$ components. Each primitive also encodes a color $\boldsymbol{c}$ value via spherical harmonics and an opacity $o$ value utilized in alpha-compositing. During rendering, 3D Gaussians are projected into 2D Gaussians based on the camera pose, sorted by their z-depths, and the final pixel color is accumulated using volumetric alpha blending:
\begin{align}
  {\bf \hat{C}}  & = \sum_{i \in N} {\bf{c}}_{i}\alpha_i \prod_{j = 1}^{i - 1} (1- \alpha_j)
\end{align}
where $i$ is the index of a Gaussian, $\alpha_i$ is calculated by multiplying 2D Gaussian's contribution with its opacity $o$.

Depth maps can be estimated from the Gaussian scene using the same discrete volume rendering equation ~\cite{turkulainen2024dn}:
 \begin{align}
  {\hat{D}}  & = \frac{\sum_{i} {{d}}_{i}\alpha_i \prod_{j = 1}^{i - 1} (1- \alpha_j)} {\sum_{i}{\alpha_i \prod_{j = 1}^{i - 1} (1- \alpha_j))}}
 \end{align}
where $\textit{d}_{i}$ is a Gaussian's z-depth in view space, and the denominator normalizes alpha estimates. However, as noted in prior research \cite{radl2024stopthepop}, this is just an approximation for per-pixel depth estimates.

To extend 3DGS's geometric accuracy, 2DGS~\cite{Huang2DGS2024} represents the scene with flat 2D Gaussians that better capture surfaces. Each primitive is composed of center points, two principal tangential vectors $t_u$ and $t_v$, and a scaling vector $\mathbf{S}=\left(s_u, s_v\right)$ to control the variance of the 2D Gaussian. With this formulation, 2DGS ensures that per-pixel depth estimates are exact and calculated explicitly through a ray-plane intersection.

In this work, we demonstrate a regularization strategy applicable to both 3D and 2D Gaussian variants by carefully supervising Gaussian positions in 3D space with sensor depth measurements and normal estimates from pretrained networks, while mitigating uncertainties in the priors. Our method serves as a plug-in module for all Gaussian-based representations to enhance geometry performance in real-world indoor reconstruction.

\begin{figure*}
    \centering
    \vspace{-0.2cm}
    \begin{tikzpicture}
        \node[anchor=south west, inner sep=0] (image) at (0,0) {\includegraphics[width=0.95\linewidth]{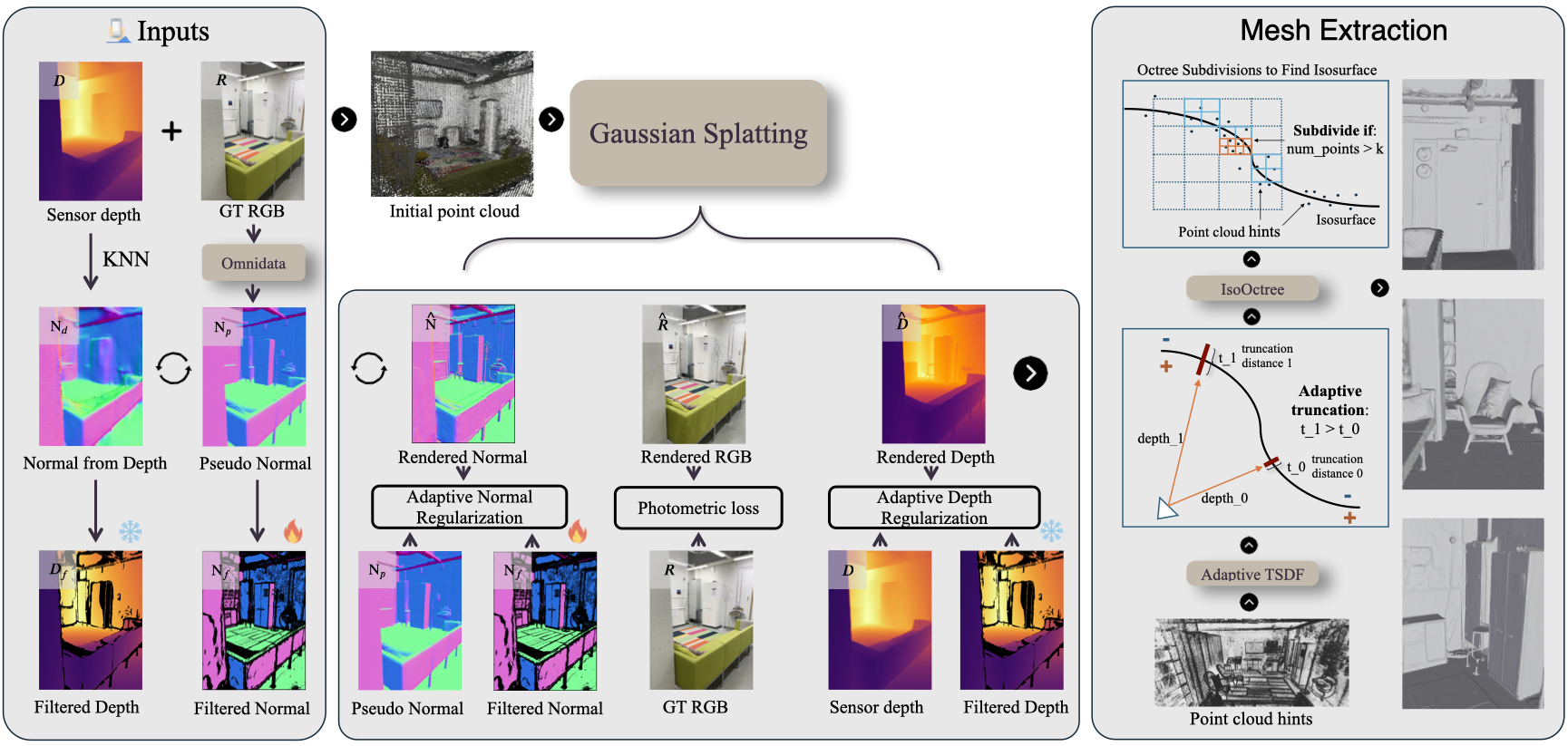}};
    \end{tikzpicture}
\caption{\textbf{Pipeline Overview}. Our approach leverages geometric consistency between normals derived from raw sensor depths and those predicted by a pretrained model to filter out noisy sensor depth data. Likewise, we compare rendered normals from a Gaussian scene with pseudo-normal estimates to dynamically filter uncertainties in normal supervision during optimization. Our adaptive depth and normal regularization terms assist various Gaussian-based frameworks in accurately reconstructing the underlying scene. Furthermore, we propose an adaptive TSDF and octree-based Marching Cubes meshing strategy enabling the extraction of smoother and more geometrically detailed meshes.}
    \label{fig:pipeline}
    \vspace{-0.6cm}
\end{figure*}

\section{Method}
\label{sec:method}
Our method consists of two adaptive supervision strategies for Gaussian Splatting-based methods that effectively combine supervision signals from geometric priors obtained from mobile devices and monocular networks. An overview of the proposed approach is illustrated in~\figref{fig:pipeline}. We first predict normal estimates from a pretrained monocular estimation model ~\citep{eftekhar2021omnidata} for input RGB images captured with a mobile device. Next, in~\secref{sec:method_depth}, we develop a depth regularization strategy that filters noisy sensor depth readings based on a normal consistency criterion. In~\secref{sec:method_normal}, we carefully utilize the pretrained monocular normal estimates for normal supervision, mitigating regularization in cases where the pretrained estimates -- due to multi-view inconsistencies or other inaccuracies -- deviate significantly from the normal estimates derived from the geometry of the optimized Gaussian scene. We describe the overall optimization process in~\secref{sec:method_optimization}. Lastly, in~\secref{sec:method_isooctree}, we propose a novel octree-based mesh extraction method that enhances surface quality and detail preservation compared to previous approaches.

\subsection{Regularization with Depth Normal Consistency}
\label{sec:method_depth}
We observe that depth sensor readings from mobile phones are relatively accurate for flat surfaces but tend to be bad for edges and far away objects. Instead, normal maps predicted from pretrained monocular models have clear object boundaries, which can serve as guidance for depth filtering. We propose an adaptive depth regularization method based on the consistency of normals derived from noisy depth images and those from pretrained networks. We coin this method as Depth Normal Consistency (DNC).

To derive robust normal estimates from noisy depth maps, we backproject depths $D(x,y)$ into world coordinates, and determine the K-Nearest Neighbors (KNN)~\cite{cover1967nearest} in world coordinates to each depth value (we set k=200). Then, a robust normal estimate $\mathbf{N}_{d}$ is generated per coordinate as the maximum eigenvector corresponding to the maximum eigenvalue of the eigenvalue decomposition of the covariance matrix determined from these K-Nearest points and their center.
% by comparing the normal orientation consistency between normal rendered from depth $N_{d}$ with K-Nearest Neighbors (KNN)~\cite{cover1967nearest} and normal $N_{M}$ predicted from the pre-train model. 

% Given per-frame $D(u,v)$ and $K$, we get point cloud $P$ from backprojection:
% \begin{equation}
% P(u,v)=D(u, v) \cdot K^{-1}\left[\begin{array}{l}
% u \\
% v \\
% 1
% \end{array}\right]
% \end{equation}
% For each point $P_i$ in the point cloud $\left\{P_i\right\}_{i=1}^N$, we find its $k$-nearest neighbors $\left\{P_{i_j}\right\}_{j=1}^k$.
% We then estimate its covariance matrix:
% \begin{equation}
% C_i=\frac{1}{k} \sum_{j=1}^k\left(P_{i_j}-\bar{P}_i\right)\left(P_{i_j}-\bar{P}_i\right)^T
% \end{equation}
% where 
% \begin{equation}
% \bar{P}_i=\frac{1}{k} \sum_{j=1}^k P_{i_j}
% \end{equation}
% The normal of $P_i$ is obtained by eigenvalue decomposition of $C_i$:
% \begin{equation}
% C_i=U_i \Lambda_i U_i^T
% \end{equation}
% $\mathbf{N}_i$ is the eigenvector $U_i$ associated with the smallest eigenvalue.
% After getting $\left\{\mathbf{N}_i\right\}_{i=1}^N$, we then backproject each point into depth map:
% \begin{equation}
% \mathbf{N}_{d}(u, v)=K\left[\begin{array}{l}
% \mathbf{N}_{x_c, i} \\
% \mathbf{N}_{y_c, i} \\
% \mathbf{N}_{z_c, i}
% \end{array}\right]
% \end{equation}
% where $x_c,y_c,z_c$ indicate the point position in camera coordination. We transfer $\mathbf{N}_d (u,v)$ to pixel coordination to get $\mathbf{N}_d$.

To filter inaccurate depth estimates, we check the orientation consistency between $\mathbf{N}_{d}$ and $\mathbf{N}_p$ generated from pre-train model with an angle threshold $\tau_d$ for filtering:
\begin{equation}
\label{eq:depth_filter}
D_{f}= \begin{cases}0 & \text{if}\ \theta_d > \tau_d \\ D& \text{otherwise} \end{cases}
\end{equation}
where
\begin{equation}
\label{theta}
\theta_d=\arccos \left(\frac{\mathbf{N}_d \cdot \mathbf{N}_p}{\left\|\mathbf{N}_d \right\|\|\mathbf{N}_p\|}\right)
\end{equation}

During training, we first regularize rendered depth estimates $\hat{D}$ with noisy sensor depth readings and then further refine depth estimates using the DNC filtered depths $D_f$:
\begin{equation}
\label{eq:depth_loss}
    \mathcal{L}_D = \begin{cases} \| \hat{D} -D\|_1 & \text{when}\ \text{step} < T_d \\ \| \hat{D} -D_f\|_1 & \text{otherwise} \end{cases} 
\end{equation}

\subsection{Adaptive Normal Regularization}
\label{sec:method_normal}
Similar to sensor depth readings, normals predicted from pretrained models $\mathbf{N}_p$ also suffer from poor multi-view consistency and precision. To alleviate the influence raised by wrong predictions, we designed an Adaptive Normal Regularization (ANR) strategy to mitigate normal supervision in uncertain regions that hard to optimize. Rendered normals $\mathbf{\hat{N}}$ derived from the Gaussian scene are supervised with the following adaptive normal loss:
\begin{equation}
\label{eq:normal_loss}
    \mathcal{L}_\mathbf{N} = \begin{cases} \| \mathbf{\hat{N}} -\mathbf{N}_p\|_1 & \text{when}\ \text{step} < T_n \\ \| \mathbf{\hat{N}} -\mathbf{N}_{f}\|_1 & \text{otherwise} \end{cases} 
\end{equation}
where $\mathbf{N}_{f}$ are filtered normals calculated by comparing the angle difference between $\mathbf{\hat{N}}$ and $\mathbf{N}_p$ with a threshold $\tau_{N}$.
\begin{equation}
\label{eq:normal_filter}
\mathbf{N}_{f}= \begin{cases}0 & \text{if}\ \theta_n > \tau_N \\ \mathbf{N}_{p} & \text{otherwise} \end{cases}
\end{equation}
where $\theta_i$ is the angle difference between $\mathbf{\hat{N}}$ and $\mathbf{N}_{p}$ like ~\cref{theta}.
%\begin{equation}
% \theta_i = \arccos \left( \frac{\mathbf{\hat{N}} \cdot \mathbf{N}_{p}}{\| \mathbf{\hat{N}} \| \| \mathbf{N}_{p} \|} \right)
% \end{equation}
The ANR strategy is designed to first regularize Gaussian normals using the fully pre-trained normals $\mathbf{N}_{p}$, and subsequently relax the training by relying only on the multi-view consistent and more reliable filtered normal $\mathbf{N}_{f}$. 

To allow the gradients from the normal loss during optimization to directly influence the Gaussian geometry, $\mathbf{\hat{N}}$ is estimated from rendered depth maps as in~\cite{Huang2DGS2024}:
\begin{equation}
\mathbf{\hat{N}}(x, y)=\frac{\nabla_x D(x,y) \times \nabla_y D(x,y)}{\left|\nabla_x D(x,y) \times \nabla_y D(x,y)\right|}
\end{equation}

\subsection{Optimization}
\label{sec:method_optimization}
Our DNC and ANR regularization terms can be adapted to several Gaussian Splatting based frameworks, including 3DGS and 2DGS methods. Take 3DGS~\cite{kerbl20233d} as an example, the final optimization loss is expressed as:
\begin{equation}
\mathcal{L} = \mathcal{L}_\text{color} + \lambda_d \mathcal{L}_{D} + \lambda_n \mathcal{L}_{\mathbf{N}}
\end{equation}
where $\mathcal{L}_\text{color}$ is the original RGB supervision loss containing a $\mathcal{L}_{1}$ and D-SSIM terms ~\cite{kerbl20233d}.

\subsection{Mesh Extraction}
\label{sec:method_isooctree}

Most prior works~\cite{guedon2023sugar, Huang2DGS2024, Yu2022MonoSDF, Izadi2011KinectFusionR3} utilize some form of Marching Cubes~\cite{lorensen1998marching} as a final processing step to extract a mesh. This also includes TSDF-based approaches offered in open-source tools like Open3D~\cite{Zhou2018}. However, due to the uniform grid discretization in naive Marching Cubes, recovering fine details in large scenes becomes computationally infeasible, as the number of evaluated points is proportional to $V/h^3$ where $V$ is the volume of the scene's bounding box and $h$ is the voxel size, which determines the finest level of detail. Alternative methods like IsoOctree~\cite{kazhdan2007unconstrained} utilize a hierarchical coarse-to-fine meshing strategy, where finer details can be preserved with adaptive grid divisions. In this work, we investigate adapting a hierarchical meshing algorithm in the context of Gaussian Splatting.

We build upon previous work by integrating a TSDF-like volume from multiple backprojected rendered depths. However, our approach introduces a key difference: we adjust the truncation distance based on the current depth value. For closer objects with small depth values, the truncation distance is reduced, whereas distant points, which tend to be less reliable, have larger truncation distances. The truncation distance follows a simple linear relationship with depth. Additionally, unlike traditional TSDF methods, we incorporate rendered normal maps alongside depth maps to filter and weight the contribution of each backprojected frame based on the consistency of their normal orientations. Our depth-aware truncated TSDF method results in a TSDF volume with a dynamically varying truncation factor.

Next, we apply the IsoOctree meshing algorithm~\cite{isooctree}, which starts with a uniform grid and progressively subdivides the volume into finer regions based on a user-defined function that determines which octree nodes to expand. To achieve this, we employ a \emph{point cloud hint}: we back-project our output depth maps from all training images into a point cloud and expand a voxel of width $h$ if it contains at least $N_{\rm e} = 50$ points within a radius $h$ from its center. This choice is based on the assumption that the appropriate level of detail in the reconstruction is proportional to the density of the point cloud, which correlates with the distance to the cameras and the number of camera views covering a particular region. To produce a smooth and artifact-free mesh, the curvature of the underlying isosurface should also follow a similar pattern: it should be low in regions of low point cloud density. To satisfy this property, our TSDF, which defines the isosurface, also employs a camera-distance-based heuristic. In~\tabref{tab:ab_optimize} and~\figref{fig:ablation_mesh}, we show that the underlying geometry from an optimized Gaussian scene can be further refined with this IsoOctree-based method. This results in a smoother surface compared to the baseline TSDF method, which lacks both subdivision and truncation distance modulation.

\begin{table*}[th!]
\vspace{-0em}
\caption{\textbf{Mesh reconstruction evaluation on MuSHRoom}. The mesh metrics are averaged over 6 scenes: "coffee\textunderscore room", "honka", "kokko", "sauna", "computer", and "vr\textunderscore room". The best results from each category are marked with \textbf{bold}. We also report Gaussian numbers after training on the "vr\textunderscore room" scene in the last column.}
\vspace{-1em}
\label{tab:main-table-mesh-mushroom}
  \centering\scriptsize
  \setlength{\tabcolsep}{2.2pt}
  \renewcommand*{\arraystretch}{1.05}
\begin{tabular}{clccccccccc}
\toprule
 & Methods & & Sensor Depth & \multicolumn{1}{c}{Meshing Algorithm} & Accuracy~$\downarrow$ & Completion~$\downarrow$  & Chamfer-$L_1$~$\downarrow$ & Normal Consistency~$\uparrow$  & F-score~$\uparrow$ & Num (M) \\
\midrule
 & Volumetric Fusion \cite{curless1996volumetric} & & \textcolor{teal}{$\checkmark$} & TSDF & .0478 & .0473 & .0476 & .7816 & .8064 & $-$ \\
\midrule 
\multirow{3}{*}{\rotatebox[origin=c]{90}{Implicit}} & Nerfacto \cite{nerfstudio} & \multirow[t]{2}{*}{NeRF} & $\textcolor{red}{-}$ & Poisson & .0430 & .0578 & .0504 & .7822 & .7212 & $-$ \\
& Depth-Nerfacto~\cite{nerfstudio} & & \textcolor{teal}{$\checkmark$} & Poisson & .0447 & .0557 & .0502 & .7614 & .6966 & $-$\\

& MonoSDF \cite{Yu2022MonoSDF} & SDF & \textcolor{teal}{$\checkmark$} & Marching-Cubes & \textbf{.0310} & \textbf{.0190} & \textbf{.0250} & \textbf{.8846} & \textbf{.9211} & $-$ \\
\midrule 
\multirow{9}{*}{\rotatebox[origin=c]{90}{Explicit}} & 
3DGS \cite{kerbl20233d} & \multirow[t]{3}{*}{3DGS} & $\textcolor{red}{-}$ & TSDF & .0929 & .0830 & .0880 & .6908 & .4228 & 5.0 \\
& SuGaR \cite{guedon2023sugar} & & $\textcolor{red}{-}$ & Poisson+IBR & \textbf{.0656} & \textbf{.0583} & \textbf{.0620} & \textbf{.8031} & \textbf{.6378} & 0.7 \\
& GOF~\cite{yu2024gaussian} & & $\textcolor{red}{-}$ & Tetrahedral & .1452 & .1102 & .1277 & .6839 & .4515 & 3.3 \\ \cline{2-11}
& Splatfacto~\cite{nerfstudio} & \multirow[t]{4}{*}{Splatfacto} & $\textcolor{red}{-}$ & Poisson & .0749 & .0555 & .0652 & .7727 & .5835 & 1.18  \\
& DN-Splatter~\cite{turkulainen2024dn} & & \textcolor{teal}{$\checkmark$}  & Poisson & \textbf{.0239} & .0194 & .0216 & \textbf{.8822} & .9243 & 1.18 \\
& DN-Splatter~\cite{turkulainen2024dn} & &  \textcolor{teal}{$\checkmark$}  & TSDF &  .0256&  .0174&  .0215&  .8390 & .9381 & 1.18 \\
\rowcolor{gray!10}
& Splatfacto~\cite{nerfstudio} + Ours  & & \textcolor{teal}{$\checkmark$}  & TSDF &  .0253 &  \textbf{.0165}&  \textbf{.0209}&  .8328&  \textbf{.9433}& 1.11 \\ \cline{2-11}
& 2DGS~\cite{Huang2DGS2024}& \multirow[t]{3}{*}{2DGS} & $\textcolor{red}{-}$ & TSDF & .0731 & .0642 & .0687 & .8008 & .6039 & 2.6 \\
\rowcolor{gray!10}
& 2DGS~\cite{Huang2DGS2024} + Ours & & \textcolor{teal}{$\checkmark$} & TSDF & .0286 & .0228 & .0257 & \textbf{.8804} & .9053 & 2.5 \\
\rowcolor{gray!10}
& 2DGS~\cite{Huang2DGS2024} + Ours & & \textcolor{teal}{$\checkmark$} & SDF + IsoOctree (Ours) & \textbf{.0249} & \textbf{.0210} & \textbf{.0229} & .8754 & \textbf{.9146}& 2.5 \\
\bottomrule
\end{tabular}
\vspace{-1em}
\end{table*}

% \usepackage{multirow}
\begin{table*}[h!]
\vspace{0em}
\caption{\textbf{Novel view synthesis evaluation} on the MuSHRoom dataset. The reported results are based on two distinct evaluation datasets: a test set obtained by uniformly sampling every 10 frames within the same training sequence, and a test set obtained from a completely different camera trajectory with no overlap with the training sequence. Results are averaged over 6 scenes.}
\vspace{-0.7em}
\label{tab:nvs}
  \centering\scriptsize
  \setlength{\tabcolsep}{10pt}
  \renewcommand*{\arraystretch}{1.05}
\begin{tabular}{clccccc|ccc}
\toprule
& \multirow{2}{*}{Method} & & \multirow{2}{*}{Sensor Depth} & \multicolumn{3}{c|}{Test  within a sequence} & \multicolumn{3}{c}{Test with a different sequence} \\ 
&                       & &                              & PSNR $\uparrow$ & SSIM $\uparrow$ & LPIPS $\downarrow$ &PSNR $\uparrow$ &  SSIM $\uparrow$  & LPIPS $\downarrow$    \\ \midrule
\multirow{3}{*}{\rotatebox[origin=c]{90}{Implicit}} & Nerfacto~\cite{nerfstudio} & \multirow[t]{2}{*}{NeRF} &  $\textcolor{red}{-}$           & 20.86        & .7859       & .2321       & 20.52          & .7705         & .2560         \\
& Depth-Nerfacto~\cite{nerfstudio} & &  \textcolor{teal}{$\checkmark$} & 21.24        & .7832       & .2414       & 20.87          & .7682         & .2643         \\
& MonoSDF~\cite{Yu2022MonoSDF}  & SDF &  \textcolor{teal}{$\checkmark$} & 20.68        & .7357       & .3590       & 19.08          & .7132         & .3820         \\
\midrule
\multirow{7}{*}{\rotatebox[origin=c]{90}{Explicit}} & 3DGS~\cite{kerbl20233d}          & \multirow[t]{3}{*}{3DGS} &  $\textcolor{red}{-}$           & 22.65        & .8286       & .1366       & 20.19          & .7574         & .1984         \\
& SuGaR \cite{guedon2023sugar}     & &  $\textcolor{red}{-}$           & 20.52        & .7740       & .2427       & 18.18          & .7125         & .2959         \\
& GOF~\cite{yu2024gaussian}        & &  $\textcolor{red}{-}$           & 19.03        & .7845       & .2189       & 17.76          & .7216         & .2816         \\ 
\cline{2-10}
& Splatfacto~\cite{nerfstudio}     &  \multirow[t]{3}{*}{Splatfacto} & $\textcolor{red}{-}$           & 24.47        & .8465       & .1358       & 21.66          & .7887         & .1922         \\
& DN-Splatter~\cite{turkulainen2024dn} & &  \textcolor{teal}{$\checkmark$} & 24.58 & .8558 & .1293 & 21.89 & .7984 & .1797 \\
\rowcolor{gray!10}
& Splatfacto~\cite{nerfstudio} + Ours & & \textcolor{teal}{$\checkmark$} & \textbf{24.83} & \textbf{.8589} & \textbf{.1129} & \textbf{22.24} & \textbf{.8054} & \textbf{.1589} \\
\cline{2-10}
& 2DGS~\cite{Huang2DGS2024} & \multirow[t]{2}{*}{2DGS} &  $\textcolor{red}{-}$           & 22.52        & .8185       & .1773       & 20.04          & .7587         & .2292         \\ 
\rowcolor{gray!10}
& 2DGS~\cite{Huang2DGS2024} + Ours & &  \textcolor{teal}{$\checkmark$}           & \textbf{23.06}       & \textbf{.8263 }      & \textbf{.1650}       & \textbf{20.97}          & \textbf{.7727}         & \textbf{.2060}         \\ 
\bottomrule
\end{tabular}
\vspace{-1em}
\end{table*}

\section{Experiments}
\label{sec:exp}
In this section, we demonstrate the effectiveness of our adaptive depth and normal regularization strategy and our proposed IsoOctree meshing method. We evaluate mesh reconstruction performance and novel-view synthesis quality.

\boldparagraph{Datasets.}
We focus on real-world indoor scenes captured using a mobile device. We select two datasets containing iPhone captures with depth data: a) MuSHRoom~\cite{ren2023mushroom} dataset: a real-world indoor dataset with different trajectories for training and evaluation; b) ScanNet++~\cite{yeshwanthliu2023scannetpp} dataset: a large scale real-world indoor dataset with high fidelity 3D geometry and RGB data.

\boldparagraph{Baselines.}
We compared our method to the following baselines: a) Traditional 3D reconstruction method Volumetric Fusion~\cite{curless1996volumetric}.  b) state-of-the-art NeRF-based method Nerfacto~\cite{nerfstudio}; c) its depth regularized version Depth-Nerfacto with a depth supervision loss similar to DS-NeRF \cite{kangle2021dsnerf}; d) MonoSDF~\cite{Yu2022MonoSDF} for SDF-based implicit surface reconstruction; e) original 3DGS~\cite{kerbl20233d} method and its advanced open-source reimplementation Splatfacto \cite{nerfstudio}; f) the following Gaussian based models focusing on mesh reconstruction: SuGaR~\cite{guedon2023sugar}, 2DGS~\cite{Huang2DGS2024}, and GOF~\cite{yu2024gaussian}; g) and lastly, DN-Splatter~\cite{turkulainen2024dn}, a 3DGS-based method that also utilizes depth and normal supervision, similar to our work.

\boldparagraph{Evaluation metrics.}
For mesh reconstruction evaluation, we follow the evaluation protocol from ~\cite{ren2023mushroom,wang2022go} and report Accuracy (Acc.), Completion (Comp.), Chamfer-$L_1$ distance (C-$L_1$), Normal Consistency (NC), and F-scores (F1) with a threshold of 5cm. For novel-view synthesis, we report PSNR, SSIM, and LPIPS metrics.

\input{author-kit-3DV2025-v1/tables/main_figure_mesh}

\boldparagraph{Implementation details.}
We implement our method using two recent open-source Gaussian splatting frameworks 2DGS~\cite{Huang2DGS2024} and Splatfacto~\cite{nerfstudio} (a 3DGS re-implementation). For all TSDF based meshing baselines, we use the open-source implementation from Open3D \cite{Zhou2018} similar to prior work \cite{Huang2DGS2024} with truncation distance of 0.03, depth truncation 10, and voxel size of 0.01. More settings and implementation details can be seen in the supplementary materials.

\subsection{3D Reconstruction Evaluation}

We evaluate mesh reconstruction performance against baselines for the MuSHRoom in ~\tabref{tab:main-table-mesh-mushroom} and~\figref{fig:different-method-cmp}, and ScanNet++ datasets~\footnote{Please refer to the table in the supplementary materials}. We demonstrate that our method can outperform the traditional volumetric fusion. Additionally, we showcase the performance of our proposed IsoOctree extraction method within the 2DGS-based framework, highlighting its superior surface creation quality. It is important to note that real-world indoor room reconstruction remains a significant challenge for existing Gaussian-based frameworks that rely solely on photometric supervision. Adding sensor depth regularization greatly enhances reconstruction quality in ambiguous, textureless regions. We also compare our method to the recent DN-Splatter~\cite{turkulainen2024dn} method, which utilizes sensor depth and normal priors for regularization. Our results demonstrate that the novel adaptive depth and normal regularization terms we propose (also showcased in the ablation study~\tabref{tab:ab_optimize}) improve mesh quality by effectively filtering out uncertain priors. Additionally, our strategy decreases the overall Gaussian count in a scene while providing equal or better novel-view synthesis results (\cf ~\tabref{tab:nvs}).

\subsection{Novel View Synthesis}
\begin{figure*}[t]
    \centering
    \begin{tikzpicture}
        \node [inner sep=0pt,outer sep=0pt,clip,rounded corners=2pt] at (0,0) {\includegraphics[width=0.9\linewidth]{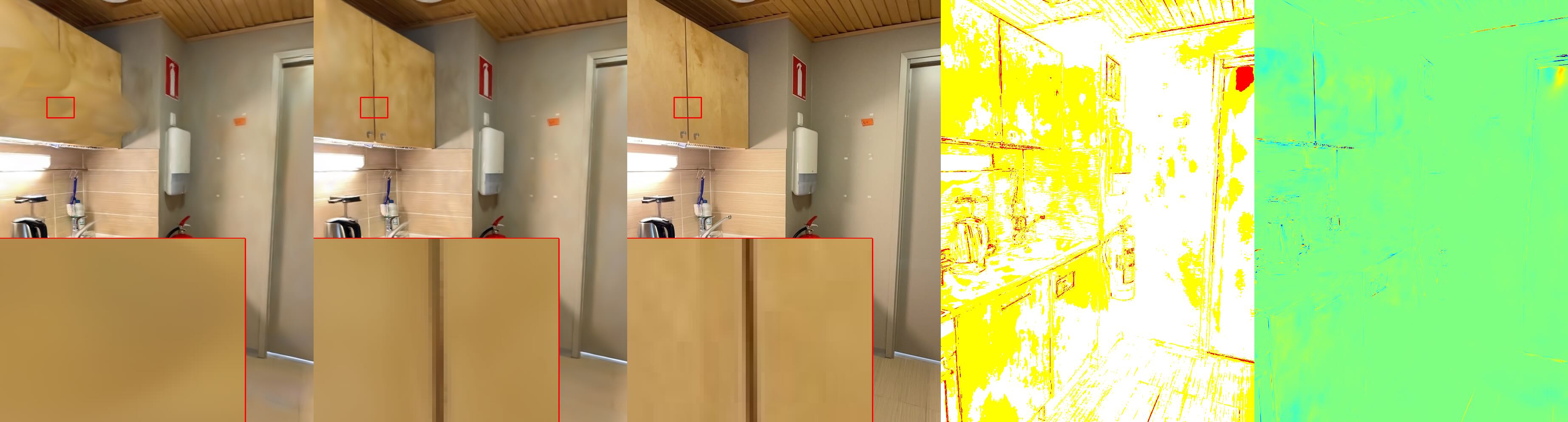}};
    \end{tikzpicture}
    \begin{tikzpicture}
        \node at (-6.3, -2.4) {\small \textbf{2DGS}\cite{Huang2DGS2024}};
        \node at (-3, -2.4) {\small \textbf{2DGS + Ours}};
        \node at (0, -2.4) {\small \textbf{Reference}};
        \node at (3.3, -2.4) {\small \textbf{$l_2$ contrib. }};
        \node at (6.5, -2.4) {\small \textbf{$\Delta l_2$}};
    \end{tikzpicture}
    \vspace{-1em}
    \caption{Novel view synthesis comparisons on the MuSHRoom dataset. From left to right: 2DGS \cite{Huang2DGS2024} baseline, 2DGS with our proposed DNC and ANR optimization strategies, reference evaluation image,  $l_2$ error contributions, 2DGS + Ours (red: $30\%$, yellow: $60\%$, white: $10\%$); $l_2$ error differences 2DGS + Ours vs 2DGS (red: higher error, blue: lower error) by comparing the two rendered images. }
    \label{fig:nvs_main}
    \vspace{-0.4cm}
\end{figure*}

We evaluate our regularization strategy on novel view synthesis metrics in~\tabref{tab:nvs}.
Rendering quality benefits from prior regularization, particularly when viewing from camera positions with less overlap with the training sequence (PSNR$+0.54/0.25$dB for the test set within the training sequence and $+0.93/0.35$dB for the test set from a wholly different camera trajectory for 2DGS/DN-Splatter). We visualize novel view synthesis examples in~\figref{fig:nvs_main}.

\subsection{Ablation Studies}
\boldparagraph{Regularization Strategy.}
We individually test the performance of our proposed optimization terms in~\tabref{tab:ab_optimize}. We observe that utilizing noisy depths significantly improves the baseline. Our more effective filtering strategy, using adaptive depth and normal supervision, further enhances meshing quality, resulting in a $2.12\%$ increase in the F-score. Additionally, the extracted mesh can be further refined using the IsoOctree meshing method. Qualitative comparisons of each optimization term are visualized in~\figref{fig:ablation_mesh}. The IsoOctree meshing method removes some of the grid-like artifacts on smooth walls that are commonly seen in Marching Cubes-based methods. Lastly, the DNC and ANR terms help preserve details for objects and reduce overall noise.
%\begin{table}[th!]
\begin{table}[t!]
\caption{\textbf{Ablation on supervision strategy and mesh performance (MuSHRoom)}. Results are averaged over three scenes.}
\label{tab:ab_optimize}
  \centering\scriptsize
  \setlength{\tabcolsep}{3pt}
  \renewcommand*{\arraystretch}{1.05}
\begin{tabular}{lccccc}
\toprule
\multicolumn{1}{c}{Input}    & Acc. $\downarrow$ & Comp. $\downarrow$ & C-$L_1$ $\downarrow$ & NC $\uparrow$ & F1 $\uparrow$       \\ \hline
2DGS \cite{Huang2DGS2024}                          & .0676 & .0605 & .0641 & .8156 & .6345 \\
+ Depth                      & .0316 & .0246 & .0281 & .8800 & .8861 \\
+ Normal                     & .0659 & .0652 & .0656 & .8678 & .6342 \\
+ Both             & .0283       &  .0247      & .0265       &   .8937     &  .8880      \\
+ Both + DNC       & .0265       & .0227       & .0246       & \textbf{.8962}       &  .9061      \\
+ Both + DNC + ANR &  .0262  & .0219       & .0241       & .8927       & .9092       \\ 
+ Both + DNC + ANR + IsoOctree &  \textbf{.0235}  &  \textbf{.0217}     &  \textbf{.0226}      &  .8888      &    \textbf{.9157}    \\ \bottomrule
\end{tabular}
\vspace{-1em}
\end{table}
\begin{table}[t!]
%\vspace{-0.2em}
  \centering\scriptsize
  \setlength{\tabcolsep}{3pt}
  \renewcommand*{\arraystretch}{1.05}
  \caption{\textbf{Ablation on monocular and sensor depth supervision} on the "vr\textunderscore room" scene from MuSHRoom. We note that monocular depth supervision greatly under-performs compared to directly using noisy sensor depth readings for indoor room reconstruction.}
\begin{tabular}{lccccc}
\toprule
Input & Acc. $\downarrow$  & Comp. $\downarrow$   &  C-$L_1$ $\downarrow$  & NC $\uparrow$  &   F1 $\uparrow$   \\ \hline
2DGS \cite{Huang2DGS2024}       & .0652          & .0673          & .0662          & .8275          & .6476          \\
+ Zoe depth~\cite{bhat2023zoedepth}   & .0448          & .0480          & .0464          & .8754          & .7224          \\
+ raw sensor depth (no filtering) & \textbf{.0216} & \textbf{.0217} & \textbf{.0216} & \textbf{.9039} & \textbf{.9051} \\ \bottomrule
\end{tabular}
\vspace{-2.2em}
\label{tab:ab-necessary-sensor-depth}
\end{table}
\input{author-kit-3DV2025-v1/tables/ablation_mesh}

\boldparagraph{Sensor Depth \vs Monocular Depth.}
In~\tabref{tab:ab-necessary-sensor-depth}, we demonstrate that sensor depths remain crucial for indoor room meshing. We compare our approach with the Zoe-Depth~\cite{bhat2023zoedepth} monocular network estimates, utilizing the recently proposed Patch-based Depth Correlation Loss~\cite{xiong2023sparsegs} for monocular depth supervision. For raw sensor depth input, we simply apply the $L_1$ loss. The quality of monocular depth reconstruction is significantly lower than that of sensor depth reconstruction because of a persistent domain gap in real-world applications.

\section{Conclusion}
\label{sec:conclusion}
In this work, we presented two regularization strategies to adaptively integrate unreliable geometric priors into Gaussian Splatting based frameworks allowing for better mesh extraction and novel-view synthesis using a mobile device. We showed how filtering noisy sensor depth readings with a normal consistency check and how mitigating uncertainties in monocular normal estimates can be used to better guide optimization. Moreover, we present a promising alternative to traditional meshing techniques using a depth adaptive TSDF and IsoOctree meshing method that can extract finer details from a Gaussian scene. Our method serves as an easy plug-in module to existing Gaussian-based frameworks and we demonstrate the effectiveness of our strategy against baseline methods with thorough experiments.

{
    \small
    \bibliographystyle{ieeenat_fullname}
    \bibliography{main}
}

\clearpage
\appendix
\setcounter{page}{1}
\maketitlesupplementary

In this supplementary material, we provide additional details regarding our AGS-Mesh optimization and the proposed Adaptive TSDF and IsoOctree meshing strategy in \cref{implementation details}. We also give further details about the meshing strategies in \cref{supp:mesh_methods}. Lastly, we present qualitative renders for mesh reconstruction and novel-view synthesis in \cref{more_exp}.

\section{Implementation Details}
\label{implementation details}
\subsection{AGS-Mesh Optimization}
We implement our regularization terms on top of the open-source implementations from 2DGS \cite{Huang2DGS2024} and DN-Splatter \cite{turkulainen2024dn}. We enable our DNR and ANR optimization terms at training iterations $T_d = 7$k (\cf\cref{eq:depth_loss}) and $T_n = 15$k (\cf\cref{eq:normal_loss}), respectively. We enable the filtered geometry prior after a certain number of steps to allow the Gaussians to be fully supervised during the initial phase and to relax the training process in later stages.  We set the angle thresholds $\tau_d$ and $\tau_N$ used for filtering inconsistent depths and normals (refer to \cf\cref{eq:depth_filter} and \cf\cref{eq:normal_filter}) to $10^\circ$. We immediately apply depth supervision at the beginning of training and enable normal regularization only after 7k iterations. The total number of training iterations is 30k. In the final optimization loss, we set $\lambda_d$ to 0.2 and $\lambda_n$ to 0.1. We use the estimated normals from Omnidata~\cite{eftekhar2021omnidata} as pre-trained normals, as they have shown to improve 3D reconstruction performance in our experiments.

\subsection{Adaptive TSDF and IsoOctree Details}
Our proposed meshing strategy consists of constructing an isofunctional inspired by TSDF approaches that is then meshed using an octree-based Marching Cubes algorithm IsoOctree.

The meshing stage takes in depth and normal renders from the Gaussian scene and camera poses. Input depths are first filtered based on a threshold that determines nearby depth similarity, if nearby depth values differ by a margin, they are filtered out. This effectively removes object edges from the depth maps and allows using linear interpolation on the remaining valid pixels. The motivation is that depth maps on the object edges are typically inaccurate and may represent a random intermediate depth value between the foreground object and the background. The normal maps are also filtered using the same mask.

We define the isofunction as
\begin{equation}
f(x) = \sum_j w_j (d_j(x) - \tilde d_j(x)),
\end{equation}
where $d_j(x)$ is the value of the depth map $j$ at the projection of point $x$ and $ \tilde d_j(x) = (x - p_j) \cdot c_z$ is the actual depth of $x$. Here $p_j$, $c_z$ are the center and principal axis of camera $j$, respectively. The sum is taken over the values where the depth map is valid and the TSDF value $d_j(x) - \hat d_j(x)$ exceeds a lower truncation distance $-\tau \cdot d_j(x)$, which depends on the projected depth. We use $\tau = 0.05$ as the relative truncation distance.
The weight in the formula is computed using a two-pass approach where we first compute a \emph{maximum weight normal} $n(x) = n_k$, $k = \argmax_j w_j'(x)$ where
\begin{equation}
    w_j' = \frac{(d_j(x) - \tilde d_j(x)) \cdot (-r_j(x) \cdot n_j(x))}{d_j(x)^2}
\end{equation}
and, on the second pass, compare the normal map value $n_j$ and camera ray direction $r_j(x) = \frac{x - p_j}{|x - p_j|}$ to $n'$ when computing the final weight $w_j$. The factor $d_j(x)^2$ in the denominator effectively down-weights observations with a larger distance to the camera, where the uncertainty of the depth map is also assumed to be the largest.

% The actual final formula for $w_j$ is a bit horrible, so perhaps the above simplified explanation is better.
% If we promise to release the code, we could ask the reader to refer to the code package for details

% --- NOTE: removed this, the above is more accurate ---
% Next, an isofunction is constructed inspired by TSDF approaches. For a sample in 3D space, it's TSDF is determined by consecutive weighting of SDFs determined by the 1-norm between the sample and backprojected depth maps per frame. A key alteration in our TSDF approach is that we modulate the maximum allowable TSDF distance based on the current frame's depth values. Larger depths have larger truncation distances and smaller depths have smaller truncation distances. This is motivated by the observation that low-resolution depth maps often have poor estimates for far away points compared to closer ones. We extend this rationale to also the optimized Gaussian scene, where depths are far away distances are most likely not reliable. 

\begin{table*}[th!]
\caption{\textbf{Mesh reconstruction evaluation on ScanNet++}. The mesh metrics are averaged over the "b20a261fdf" and "8b5caf3398" scenes. The best results from each category are marked with \textbf{bold}. Time represents training time.}
\vspace{-1em}
\label{tab:main-table-mesh-scannetpp}
  \centering\scriptsize
  \setlength{\tabcolsep}{2.2pt}
  \renewcommand*{\arraystretch}{1.05}
\begin{tabular}{clccccccccc}
\toprule
 &  Methods& & Sensor Depth & \multicolumn{1}{c}{Meshing Algorithm} & Accuracy~$\downarrow$ & Completion~$\downarrow$  & Chamfer-$L_1$~$\downarrow$ & Normal Consistency~$\uparrow$  & F-score~$\uparrow$ & Time (min) \\
\midrule
 & Volumetric Fusion \cite{curless1996volumetric} & & \textcolor{teal}{$\checkmark$} & TSDF & .0335 & .0429 & .0382 & .7372 & .8526 & 0.17 \\ \midrule 
\multirow{3}{*}{\rotatebox[origin=c]{90}{Implicit}} &  Nerfacto~\cite{nerfstudio} & \multirow[t]{2}{*}{NeRF} & $\textcolor{red}{-}$ &  Poisson & .1305 & .1484 & .1394 & .7153 & .4698 & 8.0\\
& Depth-Nerfacto~\cite{nerfstudio} & & \textcolor{teal}{$\checkmark$} & Poisson   & .0731 & .1647 & .1189 & .6848 & .5018 & 8.1\\
& MonoSDF~\cite{Yu2022MonoSDF} & SDF & \textcolor{teal}{$\checkmark$} & Marching-Cubes & \textbf{.0303} & \textbf{.0573} & \textbf{.0438} & \textbf{.8881} & \textbf{.8577} & 47.5\\
\midrule
\multirow{9}{*}{\rotatebox[origin=c]{90}{Explicit}} &  3DGS \cite{kerbl20233d} & \multirow[t]{3}{*}{3DGS} & $\textcolor{red}{-}$ & TSDF & .1795 & .1716 & .1756 & .6578 & .1719 & 14.5\\
& SuGaR~\cite{guedon2023sugar} & & $\textcolor{red}{-}$ & Poisson + IBR   & \textbf{.0940} & .1011 & \textbf{.0975} & \textbf{.7241} & \textbf{.4367} & 70\\
& GOF \cite{yu2024gaussian} & & $\textcolor{red}{-}$ & Tetrahedral & .1398 & \textbf{.0976} & .1187 & .6998 & .3239 & 142\\ \cline{2-11}
& Splatfacto~\cite{nerfstudio} & \multirow[t]{4}{*}{Splatfacto} & $\textcolor{red}{-}$ & Poisson & .1934 & .1503 & .1719 & .6741 & .1790 & 8.9 \\
& DN-Splatter~\cite{turkulainen2024dn} & & \textcolor{teal}{$\checkmark$} & Poisson &  \textbf{.0940} & .0395 & .0667 & .8316 & .7658 & 36.9\\
& DN-Splatter~\cite{turkulainen2024dn} & & \textcolor{teal}{$\checkmark$} & TSDF &   .1069&  .0251&  .0660&  \textbf{.8539}&  .8296 & 36.9\\
\rowcolor{gray!10}
& Splatfacto~\cite{nerfstudio} + Ours & & \textcolor{teal}{$\checkmark$} & TSDF & .1060 & \textbf{.0251} & \textbf{.0655} & .8506 & \textbf{.8314} & 36.9 \\ \cline{2-11}
& 2DGS~\cite{Huang2DGS2024} & \multirow[t]{3}{*}{2DGS} & $\textcolor{red}{-}$ & TSDF & .1272 & .0798 & .1035 & .7799 & .4196 & 33.5 \\
\rowcolor{gray!10}
& 2DGS~\cite{Huang2DGS2024} + Ours &   &  \textcolor{teal}{$\checkmark$} & TSDF &   \textbf{.0264} &   .0305 &   .0285 &   .9097 &  \textbf{.9030} &  40.4 \\
\rowcolor{gray!10}
& 2DGS~\cite{Huang2DGS2024} + Ours &  &  \textcolor{teal}{$\checkmark$} & SDF + IsoOctree (Ours) &  .0269 &   \textbf{.0282} &   \textbf{.0276} &  \textbf{.9139} &  .9028 &  40.4 \\

\bottomrule
\end{tabular}
\vspace{-1em}
\end{table*}

The isofunction defined above is then meshed using an IsoOctree \cite{isooctree} approach. We utilize the backprojected point cloud constructed from rendered depth maps as a \textit{point cloud hint}. The point cloud hint serves as a subdivision criteria for IsoOctree. A uniform grid is first initialized based on an AABB enclosing the point cloud hint. If a voxel contains points above a user threshold (set to 50), the voxel is subdivided into an octant. This creates a three-dimensional octree subdivision structure that contains finer levels of detail at deeper octree depths. We set the maximum octree depth to 10.

% COMMENT by Otto: IsoOctree does not use Marching Cubes at any point, it's a separate algorithm (full of nasty edge cases, which is why it has only been implemented once by the original author)
% The resulting octree is then meshed using Marching Cubes.
% \mt{TODO: explain how normal maps are used ... probably need help with this.}
% COMMENT by Otto: Normal maps are now briefly described in the above formulas

\section{Mesh Extraction Methods}
\label{supp:mesh_methods}
In this section, we provide further details on the meshing strategies shown in~\tabref{tab:main-table-mesh-scannetpp} and~\tabref{tab:main-table-mesh-mushroom}.

\boldparagraph{TSDF.}
The Truncated Signed Distance Function (TSDF) method refers to the ScalableTSDFVolume \cite{scaledtsdf} implementation from Open3D~\cite{Zhou2018}. The method accepts depth, RGB, and camera poses as input, identifies points of interest, and calculates a TSDF from input values to extract a mesh using Marching Cubes~\cite{lorensen1998marching}. We set the depth truncation distance to 10, the voxel size to 0.01, and the SDF truncation distance to 0.03 for all TSDF marked baselines in \cref{tab:main-table-mesh-scannetpp} and \cref{tab:main-table-mesh-mushroom}.

\boldparagraph{Poisson.}
Poisson refers to the screened variant of Poisson Reconstruction~\cite{screenedPoisson} used to extract a mesh from an oriented point cloud. Optimized depth and normal maps are backprojected into world coordinates to obtain oriented points. Poisson surface reconstruction is sensitive to perturbations in the oriented point cloud; therefore, noise and multi-view inconsistencies in depth maps and backprojection can lead to poor surface generation.

\boldparagraph{Poisson + IBR.}
Poisson + IBR (Image Based Rendering) refers to the optimization strategy proposed in SuGaR~\cite{guedon2023sugar}. A coarse mesh is first obtained from the Gaussian scene at 7k iterations by Poisson reconstruction from a point cloud sampled from a level set determined by the density of the Gaussian scene. The coarse mesh is then further optimized with differentiable image-based rendering (using PyTorch3D functionality) for 15k iterations to produce a refined mesh. Mesh metrics are evaluated on this refined mesh.

\boldparagraph{Tetrahedral.}
GOF~\cite{yu2024gaussian} proposed generating a 3D bounding box for each Gaussian, then establishing tetrahedral grids within these 3D bounding boxes. Marching Tetrahedra~\cite{shen2021deep} is applied to extract triangle meshes from the tetrahedral grid, using a binary search algorithm to precisely identify the level set.

\boldparagraph{SDF + IsoOctree.}
The SDF + IsoOctree method, proposed in our paper, utilizes a depth-aware truncated TSDF calculation combined with the IsoOctree meshing method. The approach can reduce the number of mesh vertices, for example, the size of the mesh extracted with TSDF is 192MB and the mesh extracted with SDF + IsoOctree is 30MB for the "vr\textunderscore room" from MuSHRoom dataset.

\section{Explanations of Benchmark selection}
We choose Splatfacto as the representative of 3DGS-based baselines as it is an advanced version of 3DGS and well-suited for indoor room reconstruction. Additionally, we implement our method on 2DGS to demonstrate its effectiveness. Although methods such as \cite{dai2024high, wang2021neus, wang2023neus2} achieve high-quality object reconstructions, they face significant challenges in indoor room reconstruction due to their high computational requirements~\cite{dai2024high} and suboptimal feature extraction performance~\cite{wang2021neus, wang2023neus2}.

\section{More Experiments}
\label{more_exp}
\subsection{Quantitative 3D Reconstruction Evaluation on ScanNet++}

We show mesh comparison quantitative results on ScanNet++ in~\tabref{tab:main-table-mesh-scannetpp}. Our method provides an overall improvement when added to baselines.

\subsection{Visualizations of DNC and ANR}
We visualize the output depth and normal maps produced by the DNC and ANR filtering terms in \figref{fig:filter_cmp}. We observe that the DNC and ANR terms successfully filter our unreliable edges and outlier depth and normal estimates, preventing them from misleading the Gaussian training process.

\subsection{Qualitative Comparision of 3D Reconstruction}
Similarly, we show additional qualitative comparisons of 3D mesh reconstruction quality on  the ScanNet++ dataset in \figref{fig:scannetpp-cmp}. Our method presents a notable improvement in smoothing flat surfaces on the extracted mesh.

\subsection{Qualitative Comparision of Novel View Synthesis}
Lastly, we compare the quality of novel view synthesis with our method along with error visualizations in \figref{fig:nvs}. We compare 2DGS with and without our DNC and ANR regularization terms with highlighted details and $l_2$ differences. We demonstrate that regularization with more accurate geometric priors not only helps mesh reconstruction, but also aids in novel view rendering, especially for removing floaters.

\section{Limitations and future work}
Our method targets 3D reconstruction using RGB sequences with sensor depth. In future work, the method could be extended to only use RGB images. The IsoOctree meshing technique we propose focuses on reducing the number of vertices and faces in the mesh while smoothing the surface. However, it does not consistently enhance the overall quality of 3D reconstructions.

\section{Acknowledgments}
We acknowledge funding from the Academy of Finland (grant No. 362409, 353139, 327911 and 353138) and support from the Wallenberg AI, Autonomous Systems and Software Program (WASP) funded by the Knut and Allice Wallenberg Foundation. MT was funded by the Finnish Center for Artificial Intelligence (FCAI). We also acknowledge CSC – IT Center for Science, Finland, for computational resources. 

\input{author-kit-3DV2025-v1/tables/supp_mesh_figs_m}
\input{author-kit-3DV2025-v1/tables/filter}
\input{author-kit-3DV2025-v1/tables/nvs_cmp}

\end{document}